\documentclass[twoside]{article}
	
\PassOptionsToPackage{numbers, compress}{natbib}
\usepackage{microtype}
\usepackage{graphicx}
\usepackage{subfig}
\usepackage{adjustbox}
\usepackage{booktabs} % for professional tables
\usepackage[utf8]{inputenc}
\usepackage{amsmath}
\usepackage{amsthm}
\usepackage{amssymb}
\usepackage{algorithm}
\usepackage{algorithmic}
\usepackage{mathtools}
\usepackage{mathrsfs}
\usepackage{bbm}
\usepackage{color}
\usepackage{cases}
\usepackage[linewidth=1pt]{mdframed}
\usepackage{amsfonts}

\usepackage{textcomp}
\def\BibTeX{{\rm B\kern-.05em{\sc i\kern-.025em b}\kern-.08em
		T\kern-.1667em\lower.7ex\hbox{E}\kern-.125emX}}

\allowdisplaybreaks

\makeatletter
\def\BState{\State\hskip-\ALG@thistlm}
\makeatother

\newtheorem{assumption}{Assumption}
\newtheorem*{thm*}{Theorem}
\newtheorem{thm}{Theorem}
\newtheorem{lemma}{Lemma}
\newtheorem*{lemma*}{Lemma}

\newtheorem*{corollary*}{Corollary}
\newtheorem{fact}{Fact}

\newtheorem*{remark*}{Remark}

\ifodd 1

\else

\fi

\ifodd 1
\newcommand{\supp}[1]{{\color{black}#1}}
\else
\newcommand{\supp}[1]{}
\fi

\ifodd 0
\newcommand{\main}[1]{{\color{black}#1}}
\else
\newcommand{\main}[1]{}
\fi

\ifodd 0
\newcommand{\revc}[1]{{\color{blue}#1}}
\else
\newcommand{\revc}[1]{#1}
\fi

\ifodd 1
\newcommand{\com}[1]{{\color{red}(C: #1)}}
\else
\newcommand{\com}[1]{}
\fi

\ifodd 0
\newcommand{\rev}[1]{{\color{blue}#1}}
\else
\newcommand{\rev}[1]{#1}
\fi

\ifodd 1
\newcommand{\dtl}[1]{{\color{red}Details:\\ #1}}
\else
\newcommand{\dtl}[1]{}
\fi

\ifodd 1
\newcommand{\alt}[1]{{\color{red}\\Alternative: #1}}
\else
\newcommand{\alt}[1]{}
\fi

\newcommand{\bs}{\boldsymbol}

\def\1{\mathbbm{1}}

\usepackage[accepted]{aistats2019}

\begin{document}
	
\runningtitle{Thompson Sampling for Combinatorial Multi-armed Bandit with Probabilistically Triggered Arms}

% \maketitle

\twocolumn[

\aistatstitle{Analysis of Thompson Sampling for Combinatorial Multi-armed Bandit with Probabilistically Triggered Arms}

\aistatsauthor{Alihan H\"uy\"uk \And Cem Tekin}

\aistatsaddress{Electrical and Electronics \\ Engineering Department \\ Bilkent University, Ankara, Turkey \And Electrical and Electronics \\ Engineering Department \\ Bilkent University, Ankara, Turkey} ]

\begin{abstract}
We analyze the regret of combinatorial Thompson sampling (CTS) for the combinatorial multi-armed bandit with probabilistically triggered arms under the semi-bandit feedback setting. We assume that the learner has access to an exact optimization oracle but does not know the expected base arm outcomes beforehand. When the expected reward function is Lipschitz continuous in the expected base arm outcomes, we derive $O(\sum_{i =1}^m \log T / (p_i \Delta_i))$ regret bound for CTS, where $m$ denotes the number of base arms, $p_i$ denotes the minimum non-zero triggering probability of base arm $i$ and $\Delta_i$ denotes the minimum suboptimality gap of base arm $i$. We also compare CTS with combinatorial upper confidence bound (CUCB) via numerical experiments \rev{on a cascading bandit problem}.
\end{abstract}

\section{INTRODUCTION}\label{sec:intro}

Multi-armed bandit (MAB) exhibits the prime example of the tradeoff between exploration and exploitation faced in many reinforcement learning problems \cite{robbins2}. In the classical MAB, at each round the learner selects an arm which yields a random reward that comes from an unknown distribution. The goal of the learner is to maximize its expected cumulative reward over all rounds by learning to select arms that yield high rewards. The learner's performance is measured by its regret with respect to an oracle policy which always selects the arm with the highest expected reward. It is shown that when the arms' rewards are independent, any uniformly good policy will incur at least logarithmic in time regret \cite{lai1}. Several classes of policies are proposed for the learner to minimize its regret. These include Thompson sampling \cite{thompson1933likelihood,agrawal2012analysis,russo2014learning} and upper confidence bound (UCB) policies \cite{lai1,agrawal1,auer}, which are shown to achieve logarithmic in time regret, and hence, are order optimal.

Combinatorial multi-armed bandit (CMAB) \cite{gai2012combinatorial,kvetonlinear,chen2013combinatorial} is an extension of MAB where the learner selects a {\em super arm} at each round, which is defined to be a subset of the {\em base arms}. Then, the learner observes and collects the reward associated with the selected super arm, and also observes the outcomes of the base arms that are in the selected super arm. This type of feedback is also called the {\em semi-bandit} feedback. For the special case when the expected reward of a super arm is a linear combination of the expected outcomes of the base arms that are in that super arm, it is shown in \cite{kvetonlinear} that a combinatorial version of UCB1 in \cite{auer} achieves $O(K m 
\log T/ \Delta)$ gap-dependent and $O( \sqrt{K m T \log T})$ gap-free regrets, where $m$ is the number of base arms, $K$ is the maximum number of base arms in a super arm, and $\Delta$ is the gap between the expected reward of the optimal super arm and the second best super arm. 

Later on, this setting is generalized to allow the expected reward of the super arm to be a more general function of the expected outcomes of the base arms that obeys certain monotonicity and bounded smoothness conditions \cite{chen2013combinatorial}. The main challenge in the general case is that the optimization problem itself is NP-hard, but an approximately optimal solution can usually be computed efficiently for many special cases \cite{Nemhauser1978}. Therefore, it is assumed that the learner has access to an approximation oracle, which can output a super arm that has expected reward that is at least $\alpha$ fraction of the optimal reward with probability at least $\beta$ when given the expected outcomes of the base arms. Thus, the regret is measured with respect to the $\alpha \beta$ fraction of the optimal reward, and it is proven that a combinatorial variant of UCB1, called CUCB, achieves $O(\sum_{i=1}^m \log T / \Delta_{i})$ regret, when the bounded smoothness function is $f(x) = \gamma x$ for some $\gamma >0$, where $\Delta_{i}$ is the minimum gap between the expected reward of the optimal super arm and the expected reward of any suboptimal super arm that contains base arm $i$.
Recently, it is shown in \cite{wang18cts} that Thompson sampling can achieve $O(\sum_{i=1}^m \log T / \Delta_{i})$ regret for the general CMAB under a Lipschitz continuity assumption on the expected reward, given that the learner has access to an exact computation oracle, which outputs an optimal super arm when given the set of expected base arm outcomes. Moreover, it is also shown that the learner cannot guarantee sublinear regret when it only has access to an approximation oracle. 

An interesting extension of CMAB is CMAB with probabilistically triggered arms (PTAs) \cite{chen} where the selected super arm probabilistically triggers a set of base arms, and the reward obtained in a round is a function of the set of probabilistically triggered base arms and their expected outcomes. For this problem, it is shown in \cite{chen} that logarithmic regret is achievable when the expected reward function has the $l_\infty$ bounded smoothness property. However, this bound depends on $1/p^*$, where $p^*$ is the minimum non-zero triggering probability. Later, it is shown in \cite{wang2017improving} that under a more strict smoothness assumption on the expected reward function, called triggering probability modulated bounded smoothness, it is possible to achieve regret which does not depend on $1/p^*$. It is also shown in this work that the dependence on $1/p^*$ is unavoidable for the general case. In another work \cite{saritac2017CMAB}, CMAB with PTAs is considered for the case when the arm triggering probabilities are all positive, and it is shown that both CUCB and CTS achieve bounded regret. 

Apart from the works mentioned above, numerous other works also tackle related online learning problems. For instance, \cite{kvetonmatroid} considers matroid MAB, which is a special case of CMAB where the super arms are given as independent sets of a matroid with base arms being the elements of the ground set, and the expected reward of a super arm is the sum of the expected outcomes of the base arms in the super arm. In addition, Thompson sampling is also analyzed for a parametric CMAB model given a prior with finite support in \cite{gopalan2014thompson}, and a contextual CMAB model with a Bayesian regret metric in \cite{wen2015efficient}. Unlike these works, we adopt the models in \cite{wang18cts} and \cite{chen}, work in a setting where there is an unknown but fixed parameter (expected outcome) vector, and analyze the expected regret. 

To sum up, in this work we analyze the (expected) regret of CTS when the learner has access to an exact computation oracle, and prove that it achieves $O(\sum_{i=1}^m \log T/(p_i \Delta_{i}))$ regret. Comparing this with the regret lower bound for CMAB with PTAs given in Theorem 3 in \cite{wang2017improving}, we also observe that our regret bound is tight. 

The rest of the paper is organized as follows. Problem formulation is given in Section \ref{sec:probform}. CTS algorithm is described in Section \ref{sec:algorithm}. Regret analysis of CTS is given in Section \ref{sec:regret}. Numerical results are given in Section \ref{sec:numerical}, and concluding remarks are given in Section \ref{sec:conc}. Additional numerical results and proofs of the lemmas that are used in the regret analysis are given in the supplemental document.

\section{PROBLEM FORMULATION}\label{sec:probform}
CMAB is a decision making problem where the learner interacts with its environment through $m$ base arms, indexed by the set $[m]:=\{1,2,...,m\}$ sequentially over rounds indexed by $t\in[T]$. In this paper, we consider the general CMAB model introduced in \cite{chen} and borrow the notation from \cite{wang18cts}. In this model, the following events take place in order in each round $t$:
\begin{itemize}
    \item The learner selects a subset of base arms, denoted by $S(t)$, which is called a super arm.
    \item $S(t)$ causes some other base arms to probabilistically trigger based on a stochastic triggering process, which results in a set of triggered base arms $S'(t)$ that contains $S(t)$.
    \item The learner obtains a reward that depends on $S'(t)$ and observes the outcomes of the arms in $S'(t)$.
\end{itemize}

Next, we describe in detail the arm outcomes, the super arms, the triggering process, and the reward. 

At each round $t$, the environment draws a random outcome vector $\pmb{X}(t):=(X_1(t),X_2(t),\ldots,X_m(t))$ from a fixed probability distribution $D$ on $[0,1]^m$ independent of the previous rounds, where $X_i(t)$ represents the outcome of base arm $i$. \rev{$D$ is unknown by the learner, but it belongs to a class of distributions ${\cal D}$ which is known by the learner.} We define the mean outcome (parameter) vector as $\pmb{\mu}:=(\mu_1,\mu_2,\ldots,\mu_m)$, where $\mu_i := \mathbb{E}_{\pmb{X} \sim D}[X_i(t)]$, and use $\pmb{\mu}_S$ to denote the projection of $\pmb{\mu}$ on $S$ for $S\subseteq[m]$.

Since CTS computes a posterior over $\pmb{\mu}$, the following assumption is made to have an efficient and simple update of the posterior distribution. 

\begin{assumption}\label{a:independence}
	The outcomes of all base arms are mutually independent, i.e., $D=D_1\times D_2\times \cdots \times D_m$.
\end{assumption}

Note that this independence assumption is correct for many applications, including the influence maximization problem with independent cascade influence propagation model \cite{kempe2003infmax}.

The learner is allowed to select $S(t)$ from a subset of $2^{[m]}$ denoted by $\mathcal{I}$, which corresponds to the set of feasible super arms. Once $S(t)$ is selected, all base arms $i\in S(t)$ are immediately triggered. These arms can trigger other base arms that are not in $S(t)$, and those arms can further trigger other base arms, and so on. At the end, a random superset $S'(t)$ of $S(t)$ is formed that consists of all triggered base arms as a result of selecting $S(t)$. \rev{We have $S'(t) \sim D^{\text{trig}}(S(t), \pmb{X}(t))$, where $D^{\text{trig}}$ is the probabilistic triggering function that describes the triggering process. For instance, in the influence maximization problem, $D^{\text{trig}}$ may correspond to the independent cascade influence propagation model defined over a given influence graph \cite{kempe2003infmax}.} The triggering process can also be described by a set of triggering probabilities. \rev{Essentially, for each $i \in [m]$ and $S \in \mathcal{I}$, $p_i^{D',S}$ denotes the probability that base arm $i$ is triggered when super arm $S$ is selected given that the arm outcome distribution is $D' \in {\cal D}$. For simplicity, we let $p_i^{S} = p_i^{D,S}$, where $D$ is the true arm outcome distribution.} Let $\tilde{S}:=\{i\in[m]:p_i^S>0\}$ be the set of all base arms that could potentially be triggered by super arm $S$, which is called the \textit{triggering set} of $S$. We have that $S(t)\subseteq S'(t)\subseteq \tilde{S}(t)\subseteq[m]$. We define $p_i:=\min_{S\in\mathcal{I}:i\in\tilde{S}}p_i^S$ as the minimum nonzero triggering probability of base arm $i$, and $p^*:=\min_{i\in[m]}p_i$ as the minimum nonzero triggering probability.

At the end of round $t$, the learner receives a reward that depends on the set of triggered arms $S'(t)$ and the outcome vector $\pmb{X}(t)$, which is denoted by $R(S'(t),\pmb{X}(t))$. For simplicity of notation, we also use $R(t) = R(S'(t),\pmb{X}(t))$ to denote the reward in round $t$. Note that whether a base arm is in the selected super arm or is triggered afterwards is not relevant in terms of the reward. We also make two other assumptions about the reward function $R$, which are standard in the CMAB literature \cite{wang18cts,chen}.

\begin{assumption}\label{a:rdependence}
The expected reward of super arm $S \in \mathcal{I}$ only depends on $S$ and the mean outcome vector $\pmb{\mu}$, i.e., there exists a function $r$ such that 
\begin{align*}
\mathbb{E}[R(t)] &= \mathbb{E}_{S'(t) \sim D^{\text{trig}}(S(t), \pmb{X}(t)),\pmb{X}(t)\sim D}[R(S'(t),\pmb{X}(t))] \\
&= r(S(t),\pmb{\mu}) .
\end{align*}
\end{assumption}

\begin{assumption}\label{a:rsmoothness} (Lipschitz continuity)
There exists a constant $B > 0$, such that for every super arm $S$ and every pair of mean outcome vectors $\pmb{\mu}$ and $\pmb{\mu'}$, $|r(S,\pmb{\mu})-r(S,\pmb{\mu'})| \leq B\|\pmb{\mu}_{\tilde{S}}-\pmb{\mu'}_{\tilde{S}}\|_1$, where $\|\cdot\|_1$ denotes the $l_1$ norm.
\end{assumption}

We consider the semi-bandit feedback model, where at the end of round $t$, the learner observes the individual outcomes of the triggered arms, denoted by $Q(S'(t),\pmb{X}(t)):=\{(i,X_i(t)):i\in S'(t)\}$. Again, for simplicity of notation, we also use $Q(t)=Q(S'(t),\pmb{X}(t))$ to denote the observation at the end of round $t$. Based on this, the only information available to the learner when choosing the super arm to select in round $t+1$ is its observation history, given as $\mathcal{F}_t:=\{(S(\tau),Q(\tau)):\tau\in[t]\}$. 

\rev{In short, the tuple $([m], {\cal I}, D, D^{\text{trig}}, R)$ constitutes a CMAB-PTA problem instance. Among the elements of this tuple only $D$ is unknown to the learner.}

In order to evaluate the performance of the learner, we define the set of optimal super arms given an $m$-dimensional parameter vector $\pmb{\theta}$ as $\text{OPT}(\pmb{\theta}):=\arg\!\max_{S\in\mathcal{I}}r(S,\pmb{\theta})$. We use $\text{OPT}:=\text{OPT}(\pmb{\mu})$ to denote the set of optimal super arms given the true mean outcome vector $\pmb{\mu}$. Based on this, we let $S^*$ to represent a specific super arm in $\arg\!\min_{S\in\text{OPT}}|\tilde{S}|$, which is the set of super arms that have triggering sets with minimum cardinality among all optimal super arms. We also let $k^* := |S^*|$ and $\tilde{k}^* := |\tilde{S}^*|$.

Next, we define the suboptimality gap due to selecting super arm $S \in {\cal I}$ as $\Delta_S:=r(S^*,\mu)-r(S,\mu)$, the maximum suboptimality gap as $\Delta_{\max}:=\max_{S\in\mathcal{I}}\Delta_S$, and the minimum suboptimality gap of base arm $i$ as $\Delta_i := \min_{S\in \mathcal{I} - \text{OPT} : i \in \tilde{S}}\Delta_S$.\footnote{If there is no such super arm $S$, let $\Delta_i = \infty$.}
The goal of the learner is to minimize the (expected) regret over the time horizon $T$, given by
\begin{align}
    \text{Reg}(T) &:= \mathbb{E}\left[ \sum_{t=1}^T(r(S^*,\pmb{\mu})-r(S(t),\pmb{\mu})) \right] \nonumber \\
    &= \mathbb{E}\left[ \sum_{t=1}^T\Delta_{S(t)} \right] \text{.} \label{eqn:regret}
\end{align}

\section{THE LEARNING ALGORITHM}\label{sec:algorithm}

We consider the CTS algorithm for CMAB with PTAs \cite{wang18cts,saritac2017CMAB} (pseudocode given in Algorithm \ref{alg:CTS}). We assume that the learner has access to an exact computation oracle, which takes as input a parameter vector $\pmb{\theta}$ \rev{and the problem structure $([m], {\cal I}, D^{\text{trig}},R)$,} and outputs a super arm, denoted by $\text{Oracle}(\pmb{\theta})$, such that $\text{Oracle}(\pmb{\theta}) \in \text{OPT}(\pmb{\theta})$. CTS keeps a Beta posterior over the mean outcome of each base arm. At the beginning of round $t$, for each base arm $i$ it draws a sample $\theta_i(t)$ from its posterior distribution. Then, it forms the parameter vector in round $t$ as $\pmb{\theta}(t) := (\theta_1(t),\ldots,\theta_m(t))$, gives it to the exact computational oracle, and selects the super arm $S(t)=\text{Oracle}(\pmb{\theta}(t))$. At the end of the round, CTS updates the posterior distributions of the triggered base arms using the observation $Q(t)$. 

\begin{algorithm}[h!]
	\caption{Combinatorial Thompson Sampling (CTS).} \label{alg:CTS}
	\begin{algorithmic}[1]
		\STATE For each base arm $i$, let $a_i=1$, $b_i=1$
		\FOR{$t=1,2,\ldots$}
		\STATE {For each base arm $i$, draw a sample $\theta_i(t)$ from Beta distribution $\beta(a_i,b_i)$; let $\pmb{\theta}(t) := (\theta_1(t),\ldots,\theta_m(t))$}
		\STATE {Select super arm $S(t)=\text{Oracle}(\pmb{\theta}(t))$, get the observation $Q(t)$}
		\FORALL{$(i,X_i)\in Q(t)$}
		\STATE $Y_i\gets 1$ with probability $X_i$, $0$ with probability $1-X_i$
		\STATE $a_i\gets a_i+Y_i$
		\STATE $b_i\gets b_i+(1-Y_i)$
		\ENDFOR
		\ENDFOR
	\end{algorithmic}
\end{algorithm}

\section{REGRET ANALYSIS}\label{sec:regret}

\subsection{Main Theorem}\label{sec:maintheorem}

The regret bound for CTS is given in the following theorem. 

\begin{thm}\label{t:gapdependent}
Under Assumptions \ref{a:independence}, \ref{a:rdependence} and \ref{a:rsmoothness}, for all $D$, the regret of CTS by round $T$ is bounded as follows:
\begin{align*}
\text{Reg}(T) &\leq \sum_{i=1}^m \max_{\revc{S \in {\cal I} - \text{OPT}} :i\in \tilde{S}} \frac{16B^2|\tilde{S}|\log T}{(1-\rho)p_i(\Delta_S - 2B(\tilde{k}^{*2}+2)\varepsilon)}\\
&\quad + \left( 3+\frac{\rev{\tilde{K}}^2}{(1-\rho)p^*\varepsilon^2}+\frac{2\mathbb{I}\{p^*<1\}}{\rho^2p^*} \right) m\Delta_{\max} \\
&\quad + \alpha\frac{8\tilde{k}^*}{p^*\varepsilon^2}\left(\frac{4}{\varepsilon^2}+1\right)^{\tilde{k}^*}\log\frac{\tilde{k}^*}{\varepsilon^2} \Delta_{\max}
\end{align*}
for all $\rho\in(0,1)$, and for all $0 < \varepsilon\leq 1/\sqrt{e}$ such that $\forall S\in\mathcal{I}-\text{OPT}$, $\Delta_S>2B(\tilde{k}^{*2}+2)\varepsilon$, where $B$ is the Lipschitz constant in Assumption \ref{a:rsmoothness}, $\alpha >0$ is a problem independent constant that is also independent of $T$, and $\rev{\tilde{K}} := \max_{S\in\mathcal{I}}|\tilde{S}|$ is the maximum triggering set size among all super arms.
\end{thm}

We compare the result of Theorem \ref{t:gapdependent} with \cite{chen}, which shows that the regret of CUCB is $O(\sum_{i\in[m]} \log T/(p_i\Delta_i))$, given an $l_{\infty}$ bounded smoothness condition on the expected reward function, when the bounded smoothness function is $f(x)=\gamma x$. When $\varepsilon$ is sufficiently small, the regret bound in Theorem \ref{t:gapdependent} is asymptotically equivalent to the regret bound for CUCB (in terms of the dependence on $T$, $p_i$ and $\Delta_i$, $i \in [m]$).

For the case with $p^*=1$ (no probabilistic triggering), the regret bound in Theorem \ref{t:gapdependent} matches with the regret bound in Theorem 1 in \cite{wang18cts} (in terms of the dependence on $T$ and $\Delta_i$, $i \in [m]$). 
As a final remark, we note that Theorem 3 in \cite{wang2017improving} shows that the $1/p_i$ term in the regret bound that multiplies the $\log T$ term is unavoidable in general.

\subsection{Preliminaries for the Proof}\label{sec:thm1prelim} 

The complement of set ${\cal S}$ is denoted by $\neg {\cal S}$ or ${\cal S}^c$. The indicator function is given as  $\mathbb{I}\{\cdot\}$. $M_i(t) := \sum_{\tau=1}^{t-1} \mathbb{I}\{i\in\tilde{S}(\tau)\}$ denotes the number of times base arm $i$ is in the triggering set of the selected super arm, i.e., it is tried to be triggered, $N_i(t) := \sum_{\tau=1}^{t-1}\mathbb{I}\{i\in S'(\tau)\}$ denotes the number of times base arm $i$ is triggered, and $\hat{\mu}_i(t) := \frac{1}{N_i(t)}\sum_{\tau:\tau<t,i\in S'(\tau)} Y_i(\tau)$ denotes the empirical mean outcome of base arm $i$ until round $t$, where $Y_i(t)$ is the Bernoulli random variable with mean $X_i(t)$ that is used for updating the posterior distribution that corresponds to base arm $i$ in CTS.

We define 
\begin{align*}
\ell(S) := \frac{2\log T}{\left(\frac{\Delta_S}{2B|\tilde{S}|}-\frac{\tilde{k}^{*2}+2}{|\tilde{S}|}\varepsilon \right)^2}
\end{align*}
as the \textit{sampling threshold} of super arm $S$, 
\begin{align}
L_i(S):=\frac{\ell(S)}{(1-\rho)p_i} \label{eqn:trialthreshold}
\end{align}
as the \textit{trial threshold} of base arm $i$ with respect to super arm $S$, and $L_i^{\max} := \max_{S \in {\cal I} - \text{OPT} :i\in\tilde{S}} L_i(S)$.

Consider an $m$-dimensional parameter vector $\pmb{\theta}$. Similar to \cite{wang18cts}, given $Z\subseteq\tilde{S}^*$, we say that the \textit{first bad event for $Z$}, denoted by $\mathcal{E}_{Z,1}(\pmb{\theta})$, holds when all $\pmb{\theta'}=(\pmb{\theta'}_{Z},\pmb{\theta}_{Z^c})$ such that $\|\pmb{\theta'}_{Z}-\pmb{\mu}_{Z}\|_{\infty}\leq\varepsilon$ satisfies the following properties:
\begin{itemize}
	\item $Z \subseteq \tilde{\text{Oracle}}(\pmb{\theta'})$.
	\item Either $\text{Oracle}(\pmb{\theta'}) \in \text{OPT}$ or $\|\pmb{\theta'}_{\tilde{\text{Oracle}}(\pmb{\theta'})}-\pmb{\mu}_{\tilde{\text{Oracle}}(\pmb{\theta'})}\|_1 > \frac{\Delta_{\text{Oracle}(\pmb{\theta'})}}{B}-(\tilde{k}^{*2}+1)\varepsilon$.
\end{itemize}

Given the same parameter vector $\pmb{\theta}$, the \textit{second bad event for $Z$} is defined as $\mathcal{E}_{Z,2}(\pmb{\theta}) := \|\pmb{\theta}_{Z}-\pmb{\mu}_{Z}\|_{\infty}>\varepsilon$.

In addition, similar to the regret analysis in \cite{wang18cts}, we will make use of the following events when bounding the regret:
\begin{align}
\mathcal{A}(t) &:= \left\{ S(t)\not\in\text{OPT} \right\} \label{eqn:A} \\
\mathcal{B}_{i,1}(t) &:= \left\{ |\hat{\mu}_i(t)-\mu_i|>\frac{\varepsilon}{|\tilde{S}(t)|} \right\} \nonumber \\
\mathcal{B}_{i,2}(t) &:= \left\{ N_i(t) \leq (1-\rho)p_iM_i(t) \right\} \label{eqn:compares1} \\
\mathcal{B}(t) &:= \left\{ \exists i\in\tilde{S}(t) : \mathcal{B}_{i,1}(t) \vee \mathcal{B}_{i,2}(t) \right\}  \label{eqn:B} \\
 \mathcal{C}(t) &:= \left\{ \exists i\in\tilde{S}(t) : |\theta_i(t)-\hat{\mu}_i(t)|>\sqrt{\frac{2\log T}{N_i(t)}} \right\} \label{eqn:C} \\
\mathcal{D}(t) &:= \left\{ \|\pmb{\theta}_{\tilde{S}(t)}(t)-\pmb{\mu}_{\tilde{S}(t)}\|_1 > \frac{\Delta_{S(t)}}{B}-(\tilde{k}^{*2}+1)\varepsilon \right\} \label{eqn:D}
\end{align}

\subsection{Regret Decomposition}\label{sec:thm1decomp} 

Using the definitions of the events given in \eqref{eqn:A}-\eqref{eqn:D}, the regret can be upper bounded as follows:
\begin{align}
\text{Reg}(T) &= \sum_{t=1}^T \mathbb{E}[\mathbb{I}\{\mathcal{A}(t)\} \times \Delta_{S(t)}] \nonumber \\
&\leq \sum_{t=1}^T \mathbb{E}[\mathbb{I}\{\mathcal{B}(t) \wedge \mathcal{A}(t)\} \times \Delta_{S(t)}] \label{eqn:decomp1} \\
&\quad + \sum_{t=1}^T \mathbb{E}[\mathbb{I}\{\mathcal{C}(t) \wedge \mathcal{A}(t)\} \times \Delta_{S(t)}] \label{eqn:decomp2} \\
&\quad + \sum_{t=1}^T \mathbb{E}[\mathbb{I}\{\neg\mathcal{B}(t) \wedge \neg\mathcal{C}(t) \wedge \mathcal{D}(t) \wedge \mathcal{A}(t)\} \nonumber \\
&\quad\quad \times \Delta_{S(t)}] \label{eqn:decomp3} \\
&\quad + \sum_{t=1}^T \mathbb{E}[\mathbb{I}\{\neg\mathcal{D}(t) \wedge \mathcal{A}(t)\} \times \Delta_{S(t)}] ~.  \label{eqn:decomposition}
\end{align}

The regret bound in Theorem \ref{t:gapdependent} is obtained by bounding the terms in the above decomposition. All events except the one specified in \eqref{eqn:decomp3} can only happen a small (finite) number of times. Every time \eqref{eqn:decomp3} happens, there must be base arms in the triggering set of the selected super arm which are ``tried'' to be triggered less than the trial threshold. These ``under-explored'' base arms are the main contributors of the regret, and their contribution depends on how many times they are ``tried''. Moreover, every time these base arms are tried, their contribution to the future regret decreases. Thus, by summing up these contributions we obtain a logarithmic bound for \eqref{eqn:decomp3}. In the proof, we will make use of the facts and lemmas that are introduced in the following section.

\subsection{Facts and Lemmas}\label{sec:thm1lemmas}

\begin{fact}\label{fact:2}
(Lemma 4 in \cite{wang18cts})
When CTS is run, the following holds for all base arms $i \in [m]$:
\begin{align*}
\Pr\left[\theta_i(t)-\hat{\mu}_i(t)>\sqrt{\frac{2\log T}{N_i(t)}}\right] &\leq \frac{1}{T} \\
\Pr\left[\hat{\mu}_i(t)-\theta_i(t)>\sqrt{\frac{2\log T}{N_i(t)}}\right] &\leq \frac{1}{T}
\end{align*}
\end{fact}

We also have the following three lemmas (proofs can be found in the supplemental document along with some additional facts that are used in the proofs).

\begin{lemma}\label{lemma:1}
When CTS is run, we have
\begin{align*}
&\mathbb{E}[|\{ t:1\leq t\leq T, i\in\tilde{S}(t), |\hat{\mu}_i(t)-\mu_i|>\varepsilon \vee \mathcal{B}_{i,2}(t) \} |] \\
&\quad 
\leq 1+\frac{1}{(1-\rho)p^*\varepsilon^2}+\frac{2\mathbb{I}\{p^*<1\}}{\rho^2p^*}
\end{align*}
for all $i \in [m]$ and $\rho \in (0,1)$.
\end{lemma}

\begin{lemma}\label{lemma:2}
Suppose that $\neg\mathcal{D}(t) \wedge \mathcal{A}(t)$ happens. Then, there exists $Z\subseteq\tilde{S}^*$ such that $Z\neq\emptyset$ and $\mathcal{E}_{Z,1}(\pmb{\theta}(t))$ holds.
\end{lemma}

\begin{lemma}\label{lemma:3}
When CTS is run, for all $Z\subseteq\tilde{S}^*$ such that $Z\neq\emptyset$, we have 
\begin{align*}
&\sum_{t=1}^T \mathbb{E}[\mathbb{I}\{ \mathcal{E}_{Z,1}(\pmb{\theta}(t)),\mathcal{E}_{Z,2}(\pmb{\theta}(t)) \}] \\
&\quad  \leq 13\alpha'_2 \cdot \frac{|Z|}{p^*} \cdot \left( \frac{2^{2|Z|+3}\log\frac{|Z|}{\varepsilon^2}}{\varepsilon^{2|Z|+2}} \right) 
\end{align*}
where $\alpha'_2$ is a problem independent constant.
\end{lemma}

\subsection{Main Proof of Theorem 1}\label{sec:thm1mainproof} 

\subsubsection{Bounding \eqref{eqn:decomp1}}

Using Lemma \ref{lemma:1}, we have
\begin{align*}
    &\sum_{t=1}^T \mathbb{E} [ \mathbb{I}\{\mathcal{B}(t) \wedge \mathcal{A}(t)\} \times \Delta_{S(t)} ] \\
    &\quad \leq \Delta_{\max} \sum_{i=1}^m \mathbb{E} \left[  \bigg| \bigg{\{} t:1 \leq t \leq T, i \in \tilde{S}(t), \right. \\
    &\quad\quad \left. |\hat{\mu}_i(t)-\mu_i|>\frac{\varepsilon}{\rev{\tilde{K}}} \vee \mathcal{B}_{i,2}(t) \bigg{\}} \bigg|  \right] \\
    &\quad \leq \left( 1+\frac{\rev{\tilde{K}}^2}{(1-\rho)p^*\varepsilon^2}+\frac{2\mathbb{I}\{p^*<1\}}{\rho^2p^*} \right) m\Delta_{\max} .
\end{align*}

\subsubsection{Bounding \eqref{eqn:decomp2}}

By Fact \ref{fact:2}, we have
\begin{align*}
    &\sum_{t=1}^T \mathbb{E} [\mathbb{I}\{\mathcal{C}(t) \wedge \mathcal{A}(t)\} \times \Delta_{S(t)}] \\
    &\quad \leq \Delta_{\max} \sum_{i=1}^m \sum_{t=1}^T 
    \Pr \left[|\theta_i(t)-\hat{\mu}_i(t)| > \sqrt{\frac{2\log T}{N_i(t)}} \right] \\
    &\quad \leq 2m\Delta_{\max} .
\end{align*}

\subsubsection{Bounding \eqref{eqn:decomp3}}

For this, we first show that event $\neg\mathcal{B}(t) \wedge \neg\mathcal{C}(t) \wedge \mathcal{D}(t) \wedge \mathcal{A}(t)$ cannot happen when $M_i(t)>L_i(S(t))$, $\forall i \in \tilde{S}(t)$.
To see this, assume that both $\neg\mathcal{B}(t) \wedge \neg\mathcal{C}(t) \wedge \mathcal{A}(t)$ and $M_i(t)>L_i(S(t))$, $\forall i \in \tilde{S}(t)$ holds. Then, we must have
\begin{align}
    &\|\pmb{\theta}_{\tilde{S}(t)}(t)-\hat{\pmb{\mu}}_{\tilde{S}(t)}(t)    \|_1 \nonumber \\
    &\quad = \sum_{i\in\tilde{S}(t)}|\theta_i(t)-\hat{\mu}_i(t)| \nonumber \\
    &\quad \leq \sum_{i\in\tilde{S}(t)}\sqrt{\frac{2\log T}{N_i(t)}} \label{eqn:decomp3_1}\\
    &\quad \leq \sum_{i\in\tilde{S}(t)}\sqrt{\frac{2\log T}{(1-\rho)p_iM_i(t)}} \label{eqn:decomp3_2}\\
    &\quad \leq \sum_{i\in\tilde{S}(t)}\sqrt{\frac{2\log T}{(1-\rho)p_iL_i(S(t))}} \nonumber \\
    &\quad = \sum_{i\in\tilde{S}(t)}\sqrt{\frac{2\log T}{\ell(S(t))}} \label{eqn:decomp3_3} \\
    &\quad = \sum_{i\in\tilde{S}(t)}\left( \frac{\Delta_{S(t)}}{2B|\tilde{S}(t)|}-\frac{\tilde{k}^{*2}+2}{|\tilde{S}(t)|}\varepsilon \right) \nonumber \\
    &\quad = |\tilde{S}(t)|\left( \frac{\Delta_{S(t)}}{2B|\tilde{S}(t)|}-\frac{\tilde{k}^{*2}+2}{|\tilde{S}(t)|}\varepsilon \right) \nonumber \\
    &\quad = \frac{\Delta_{S(t)}}{2B} - (\tilde{k}^{*2}+2)\varepsilon \nonumber 
\end{align}
where \eqref{eqn:decomp3_1} holds when $\neg\mathcal{C}(t)$ happens, \eqref{eqn:decomp3_2} holds when $\neg\mathcal{B}(t)$ happens, and \eqref{eqn:decomp3_3} holds by the definition of $L_i(S(t))$.

We also know that $\|\hat{\pmb{\mu}}_{\tilde{S}(t)}(t) - \pmb{\mu}_{\tilde{S}(t)}\|_1 \leq \varepsilon$, when $\neg\mathcal{B}(t)$ happens. Then, $\|\pmb{\theta}_{\tilde{S}(t)}(t)-\pmb{\mu}_{\tilde{S}(t)} \|_1 \leq \|\pmb{\theta}_{\tilde{S}(t)}(t)-\hat{\pmb{\mu}}_{\tilde{S}(t)}(t)\|_1 + \|\hat{\pmb{\mu}}_{\tilde{S}(t)}(t) - \pmb{\mu}_{\tilde{S}(t)}\|_1 \leq \frac{\Delta_{S(t)}}{2B} - (\tilde{k}^{*2}+1)\varepsilon < \frac{\Delta_{S(t)}}{B} - (\tilde{k}^{*2}+1)\varepsilon$, which implies that $\neg\mathcal{D}(t)$ happens. 
Thus, we conclude that when $\neg\mathcal{B}(t) \wedge \neg\mathcal{C}(t) \wedge \mathcal{D}(t) \wedge \mathcal{A}(t)$ happens, then there exists some $i \in \tilde{S}(t)$ such that $M_i(t) \leq L_i(S(t))$.
Let $S_1(t)$ be the base arms $i$ in $\tilde{S}(t)$ such that $M_i(t)>L_i(S(t))$, and $S_2(t)$ be the other base arms in $\tilde{S}(t)$. By the result above, $S_2(t) \neq \emptyset$. 
Next, we show that 
\begin{align*}
\Delta_{S(t)} \leq 2B \sum_{ i \in S_2(t) } \sqrt{\frac{2\log T}{(1-\rho)p_iM_i(t)}} .
\end{align*}
This holds since,
\begin{align*}
    &\frac{\Delta_{S(t)}}{B} - (\tilde{k}^{*2}+1)\varepsilon 
     < \sum_{i\in\tilde{S}(t)}|\theta_i(t)- \mu_i |   \\
    &\quad \leq \sum_{i\in\tilde{S}(t)}(|\theta_i(t)-\hat{\mu}_i(t)| + |\hat{\mu}_i(t)-\mu_i|) \\
    &\quad \leq \sum_{i\in S_1(t)}|\theta_i(t)-\hat{\mu}_i(t)| + \sum_{i\in S_2(t)}|\theta_i(t)-\hat{\mu}_i(t)| + \varepsilon \\
    &\quad \leq |S_1(t)|\left( \frac{\Delta_{S(t)}}{2B|\tilde{S}(t)|} - \frac{\tilde{k}^{*2}+2}{|\tilde{S}(t)|}\varepsilon \right) \\
    &\quad\quad + \sum_{i\in S_2(t)}|\theta_i(t)-\hat{\mu}_i(t)| + \varepsilon \\
    &\quad \leq \frac{\Delta_{S(t)}}{2B} - (\tilde{k}^{*2}+1)\varepsilon + \sum_{i\in S_2(t)}|\theta_i(t)-\hat{\mu}_i(t)| \\
    &\quad \leq \frac{\Delta_{S(t)}}{2B} - (\tilde{k}^{*2}+1)\varepsilon + \sum_{i\in S_2(t)} \sqrt{\frac{2\log T}{N_i(t)}} \\
    &\quad \leq \frac{\Delta_{S(t)}}{2B} - (\tilde{k}^{*2}+1)\varepsilon + \sum_{i\in S_2(t)} \sqrt{\frac{2\log T}{(1-\rho)p_iM_i(t)}} .
\end{align*}

Fix $i \in [m]$. For $w > 0$, let $\eta^i_w$ be the round for which $i \in S_2(\eta^i_w)$ and $| \{ t \leq \eta^i_w  : i \in S_2(t) \}| = w$, and $w^i(T) := | \{ t \leq T : i \in S_2(t) \}| $. We have $i\in S_2(\eta^i_w) \subseteq \tilde{S}(\eta^i_w)$ for all $w>0$, which implies that $M_i(\eta^i_{w+1}) \geq w$. Moreover, by the definition of $S_2(t)$, we know that $M_i(t) \leq L_i(S(t)) \leq L_i^{\max}$ for $i \in S_2(t)$, $t \leq T$. These two facts together imply that $w^i(T) \leq L_i^{\max}$ with probability $1$.

Consider the round $\tau^i_1$ for which $i\in\tilde{S}(t)$ for the first time, i.e., $\tau^i_1 := \min\{t:i\in\tilde{S}(t)\}$. We know that $M_i(\tau^i_1)=0\leq L_i(S)$ for all $S$, hence $i\in S_2(\tau^i_1)$. Since $\forall t<\tau^i_1$, $i\not\in\tilde{S}(t)$, and $i\not\in\tilde{S}(t)$ implies $i\not\in S_2(t)$, we conclude that $\tau^i_1=\eta^i_1$. We also observe that $\neg\mathcal{B}(t)$ cannot happen for $t \leq \tau^i_1=\eta^i_1$, since $N_i(t)>(1-\rho)p_i M_i(t)=0$ cannot be true when $N_i(t)\leq M_i(t)=0$.

Then,
\begin{align}
    &\sum_{t=1}^T \mathbb{E}[\mathbb{I}\{\neg\mathcal{B}(t) \wedge \neg\mathcal{C}(t) \wedge \mathcal{D}(t) \wedge \mathcal{A}(t)\} \times \Delta_{S(t)}] \notag \\
    &\quad \leq \mathbb{E}\left[ \sum_{t=1}^{T} \sum_{i\in S_2(t)} \mathbb{I}\{\neg\mathcal{B}(t)\} 2B\sqrt{\frac{2\log T}{(1-\rho)p_iM_i(t)}} \right] \notag \\
    &\quad = \mathbb{E}\left[  \sum_{i=1}^m \sum_{t=1}^{T} \mathbb{I}\{i\in S_2(t), \neg\mathcal{B}(t)\} \right. \notag \\ 
    &\quad\quad \times\; \left. 2B\sqrt{\frac{2\log T}{(1-\rho)p_iM_i(t)}} \right] \notag \\
    &\quad \revc{\leq} \mathbb{E}\left[  \sum_{i=1}^m \sum_{t=\eta^i_1+1}^{T} \mathbb{I}\{i\in S_2(t)\} 2B\sqrt{\frac{2\log T}{(1-\rho)p_iM_i(t)}} \right] \notag \\
    &\quad \leq \mathbb{E}\left[  \sum_{i=1}^m \sum_{w=1}^{\lfloor L_i^{\max}\rfloor} \sum_{t=\eta^i_w+1}^{\eta^i_{w+1}} \mathbb{I}\{i\in S_2(t)\} \right. \notag \\
    &\quad\quad \left. \times\; 2B\sqrt{\frac{2\log T}{(1-\rho)p_iM_i(t)}} \right] \notag \\
    &\quad = \mathbb{E}\left[  \sum_{i=1}^m \sum_{w=1}^{\lfloor L_i^{\max}\rfloor} 2B\sqrt{\frac{2\log T}{(1-\rho)p_iM_i(\eta^i_{w+1})}} \right] \notag \\
    &\quad \leq \sum_{i=1}^m \sum_{w=1}^{\lfloor L_i^{\max}\rfloor} 2B\sqrt{\frac{2\log T}{(1-\rho)p_iw}} \notag \\
    &\quad \leq \sum_{i=1}^m 4B\sqrt{\frac{2\log T}{(1-\rho)p_i}L_i^{\max}} \label{eqn:decomp3_4} 
\end{align}
where \eqref{eqn:decomp3_4} holds since $\sum_{n=1}^N\sqrt{1/n} \leq 2\sqrt{N}$.

\subsubsection{Bounding \eqref{eqn:decomposition}}

From Lemma \ref{lemma:2}, we know that
\begin{align*}
    &\sum_{t=1}^T \mathbb{E}[\mathbb{I}\{\neg\mathcal{D}(t) \wedge \mathcal{A}(t)\} \times \Delta_{S(t)}] \\
    &\quad \leq \Delta_{\max} \sum_{Z\subseteq\tilde{S}^*, Z\neq\emptyset} \left( \sum_{t=1}^T \mathbb{E}[\mathbb{I}\{ \mathcal{E}_{Z,1}(\pmb{\theta}(t)),\mathcal{E}_{Z,2}(\pmb{\theta}(t)) \}] \right) 
\end{align*}
since $\neg\mathcal{D}(t) \wedge \mathcal{A}(t)$ implies $\mathcal{E}_{Z,1}(\pmb{\theta}(t))$ for some $Z\subseteq\tilde{S}^*$, and $\mathcal{E}_{Z,1}(\pmb{\theta}(t)) \wedge \neg\mathcal{E}_{Z,2}(\pmb{\theta}(t))$ implies either $\neg\mathcal{A}(t)$ or $\mathcal{D}(t)$.

From Lemma \ref{lemma:3}, we have:
\begin{align*}
    &\sum_{Z\subseteq\tilde{S}^*, Z\neq\emptyset} \left( \sum_{t=1}^T \mathbb{E}[\mathbb{I}\{ \mathcal{E}_{Z,1}(\pmb{\theta}(t)),\mathcal{E}_{Z,2}(\pmb{\theta}(t)) \}] \right) \\
    &\quad \leq \sum_{Z\subseteq\tilde{S}^*, Z\neq\emptyset} 13\alpha'_2 \cdot \frac{|Z|}{p^*} \cdot \left( \frac{2^{2|Z|+3}\log\frac{|Z|}{\varepsilon^2}}{\varepsilon^{2|Z|+2}} \right) \\
    &\quad \leq 13\alpha'_2\frac{8\tilde{k}^*}{p^*\varepsilon^2}\log\frac{\tilde{k}^*}{\varepsilon^2} \sum_{Z\subseteq\tilde{S}^*, Z\neq\emptyset} \frac{2^{2|Z|}}{\varepsilon^{2|Z|}} \\
    &\quad \leq 13\alpha'_2\frac{8\tilde{k}^*}{p^*\varepsilon^2}\left(\frac{4}{\varepsilon^2}+1\right)^{\tilde{k}^*}\log\frac{\tilde{k}^*}{\varepsilon^2} \text{.}
\end{align*}

\subsubsection{Summing the Bounds}

The regret bound for CTS is computed by summing the bounds derived for terms \eqref{eqn:decomp1}-\eqref{eqn:decomposition} in the regret decomposition, which are given in the sections above:
\begin{align*}
\text{Reg}(T) \leq &\sum_{i=1}^m 4B\sqrt{\frac{2\log T}{(1-\rho)p_i}L_i^{\max}} \\
    &\quad + \left( 3+\frac{\rev{\tilde{K}}^2}{(1-\rho)p^*\varepsilon^2}+\frac{2\mathbb{I}\{p^*<1\}}{\rho^2p^*} \right) m\Delta_{\max} \\
    &\quad + 13\alpha'_2\frac{8\tilde{k}^*}{p^*\varepsilon^2}\left(\frac{4}{\varepsilon^2}+1\right)^{\tilde{k}^*}\log\frac{\tilde{k}^*}{\varepsilon^2}\Delta_{\max} \\
    = &\sum_{i=1}^m \max_{\revc{S \in {\cal I} - \text{OPT}} : i\in \tilde{S}} \frac{16B^2|\tilde{S}|\log T}{(1-\rho)p_i(\Delta_S - 2B(\tilde{k}^{*2}+2)\varepsilon)} \\
    &\quad + \left( 3+\frac{\rev{\tilde{K}}^2}{(1-\rho)p^*\varepsilon^2}+\frac{2\mathbb{I}\{p^*<1\}}{\rho^2p^*} \right) m\Delta_{\max} \\
    &\quad + \alpha\frac{8\tilde{k}^*}{p^*\varepsilon^2}\left(\frac{4}{\varepsilon^2}+1\right)^{\tilde{k}^*}\log\frac{\tilde{k}^*}{\varepsilon^2}\Delta_{\max} 
\end{align*}
where $\alpha:=13\alpha'_2$.

\subsection{Differences from the Analysis in \cite{wang18cts}}

Our regret analysis differs from the analysis in \cite{wang18cts} (without probabilistic triggering) in the following aspects: First of all, our bad events ${\cal E}_{Z,1}(\bs{\theta})$ and ${\cal E}_{Z,2}(\bs{\theta})$ given in Section \ref{sec:thm1prelim} are defined in terms of subsets $Z$ of $\tilde{S}^*$ rather than $S^*$. Secondly, we need to relate the number of times base arm $i$ is in the triggering set of the selected super arm ($M_i(t)$) with the number of times it is triggered ($N_i(t)$), which requires us to define events ${\cal B}_{i,2}(t)$ for $i \in [m]$ as given in \eqref{eqn:compares1}, and use them in the regret decomposition. This introduces new challanges in bounding \eqref{eqn:decomp3}, where we make use of a variable called {\em trial threshold} ($L_i(S(t))$) given in \eqref{eqn:trialthreshold} to show that \eqref{eqn:decomp3} cannot happen when $M_i(t) > L_i(S(t))$, $\forall i \in \tilde{S}(t)$. We also need to take probabilistic triggering into account when proving Lemmas \ref{lemma:1} and \ref{lemma:3}. For instance, in Lemma \ref{lemma:3}, we define a new way to count the number of times ${\cal E}_{Z,1}(\bs{\theta}) \wedge {\cal E}_{Z,2}(\bs{\theta})$ happens for all $Z \subseteq \tilde{S}^*$ such that $Z \neq \emptyset$.

\section{NUMERICAL RESULTS}\label{sec:numerical}
In this section, we compare CTS with CUCB in \cite{chen} in a cascading bandit problem \cite{kveton_cascading}, which is a special case of CMAB with PTAs. In the disjunctive form of this problem a search engine outputs a list of $K$ web pages for each of its $L$ users among a set of $R$ web pages. Then, the users examine their respective lists, and click on the first page that they find attractive. If all pages fail to attract them, they do not click on any page. The goal of the search engine is to maximize the number of clicks.

The problem can be modeled as a CMAB problem. The base arms are user-page pairs $(i,j)$, where $i\in[L]$ and $j\in[R]$. User $i$ finds page $j$ attractive independent of other users and other pages, and the probability that user $i$ finds page $j$ attractive is given as $p_{i,j}$. The super arms are $L$ lists of $K$-tuples, where each $K$-tuple represents the list of pages shown to a user. Given a super arm $S$, let $S(i,k)$ denote the $k$th page that is selected for user $i$. Then, the triggering probabilities can be written as
\begin{align*}
    p_{(i,j)}^S &= 
    \begin{cases}
        1 & \text{if } j=S(i,1) \\
        \prod_{k'=1}^{k-1} (1-p_{i,S(i,k')}) & \text{if } \exists k\neq 1 :  j=S(i,k) \\
        0 & \text{otherwise}
    \end{cases}
\end{align*}
that is we observe feedback for a top selection immediately, and observe feedback for the other selections only if all previous selections fail to attract the user. The expected reward of playing super arm $S$ can be written as
%\begin{align*}
$r(S,\pmb{p}) = \sum_{i=1}^L \left(1-\prod_{k=1}^K(1-p_{i,S(i,k)})\right)$
%\end{align*}
for which Assumption \ref{a:rsmoothness} holds when $B=1$.

\begin{figure}[!ht]
	\begin{center}	
		\includegraphics[width=\linewidth]{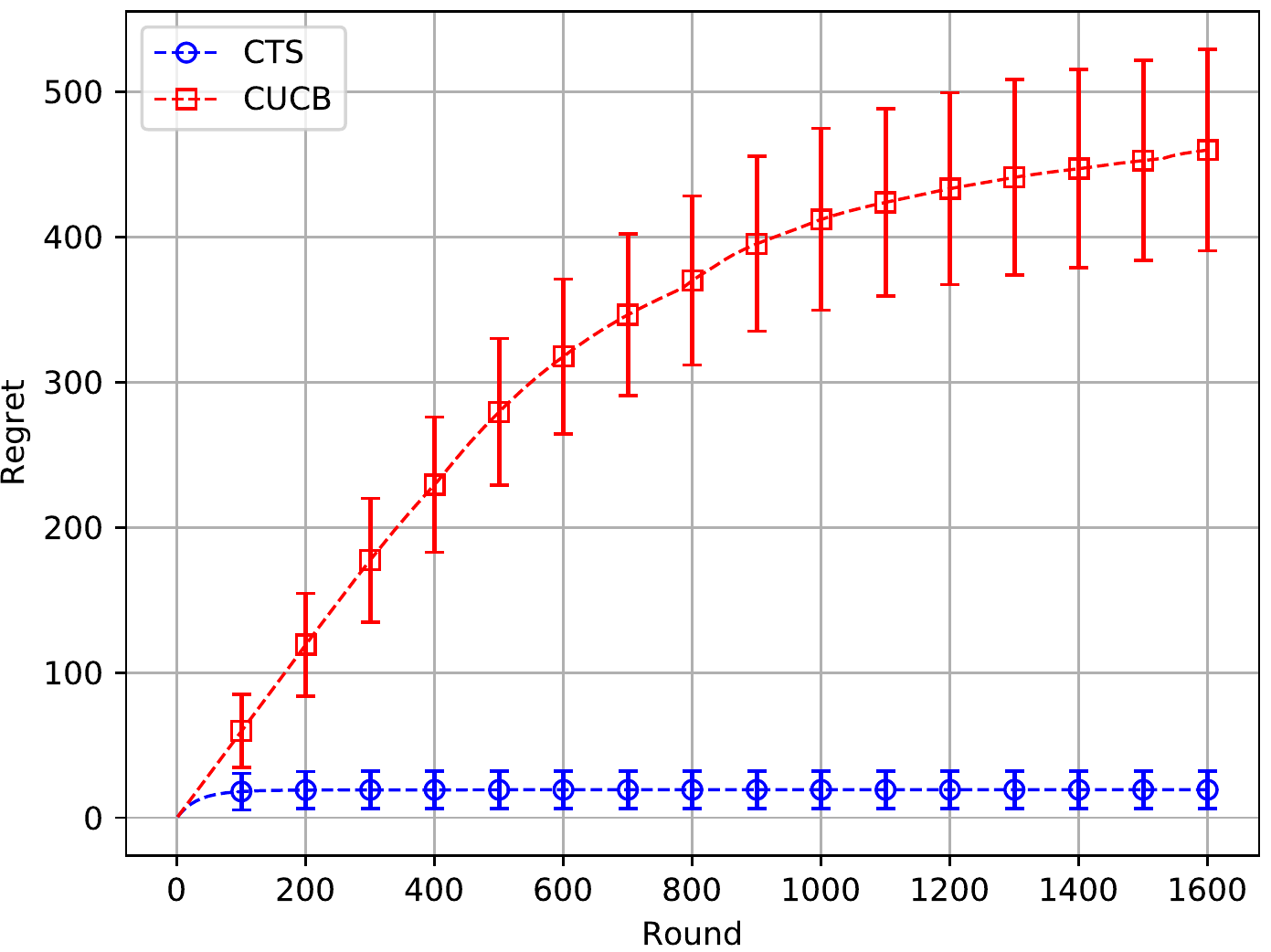}
	\end{center}
	\caption{Regrets of CTS and CUCB for the \rev{disjunctive} cascading bandit problem.}
	\label{fig:comparison}
\end{figure}

We let \rev{$L=20$, $R=100$ and $K=5$}, and generate $p_{i,j}$s by sampling uniformly at random from $[0,1]$. We run both CTS and CUCB for $1600$ rounds, and report their regrets averaged over $1000$ runs in Figure \ref{fig:comparison}, where error bars represent the standard deviation of the regret (multiplied by 10 for visibility). As expected CTS significantly outperforms CUCB. Relatively bad performance of CUCB can be explained by excessive number of explorations due to the UCBs that stay high for large number of rounds.

\begin{figure}[!ht]
	\begin{center}	
		\includegraphics[width=\linewidth]{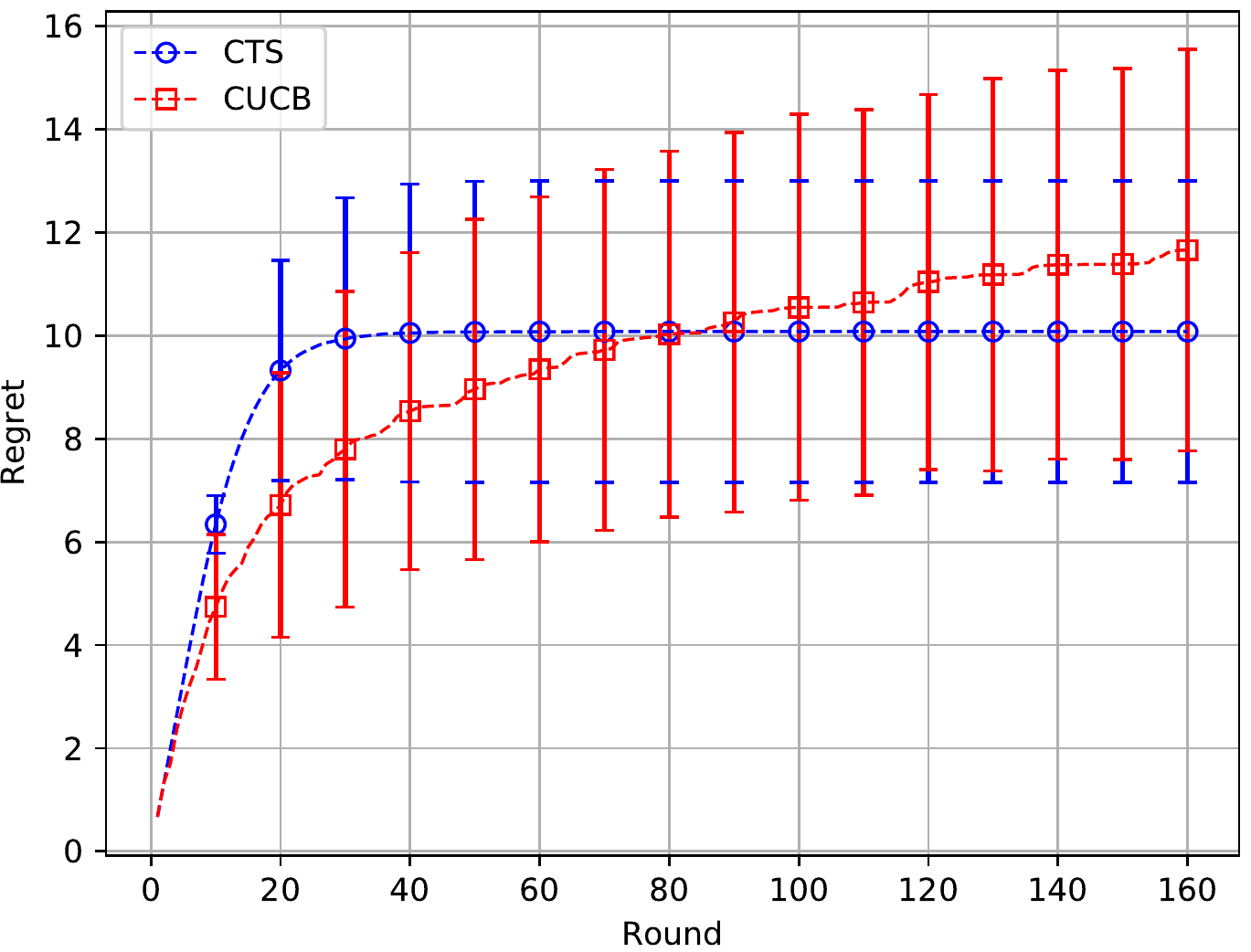}
	\end{center}
	\caption{Regrets of CTS and CUCB for the conjunctive \rev{cascading bandit} problem.}
	\label{fig:exponential}
\end{figure}

Next, we study the conjunctive analogue of the problem that we consider, where the goal of the search engine is to maximize the number of users with lists that do not contain unattractive pages, and when examining their lists, users provide feedback by reporting the first unattractive page. Formally, 
\begin{align*}
p_{(i,j)}^S &= 
\begin{cases}
1 & \text{if } j=S(i,1) \\
\prod_{k'=1}^{k-1} p_{i,S(i,k')}  &\text{if } \exists k\neq 1 : j=S(i,k)  \\
0 & \text{otherwise}
\end{cases}
\end{align*}
and
%\begin{align*}
$r(S,\pmb{p}) = \sum_{i=1}^L \prod_{k=1}^K p_{i,S(i,k)}$. 
%\end{align*}
In this setting, we let $L=1$, $R=1000000$, $K=999999$, and set all $p_{i,j}=1$ except for one, which is set to $1/3$. Essentially, we are trying to find the one page that does not certainly attract the user. We run both CTS and CUCB for $160$ rounds, and report their regrets averaged over $1000$ runs in Figure \ref{fig:exponential}, where error bars represent the standard deviation of the regret. It is observed that CUCB outperforms CTS in the first $80$ rounds and CTS outperforms CUCB after the first $80$ rounds.

This case is specifically constructed to show that CTS does not always uniformly outperform CUCB when there are many good arms but a single bad arm. For CTS to act optimally in this case, the sample from the posterior of the bad arm should be less than the samples from the posteriors of the other arms, which happens with a small probability at the beginning. CUCB is good at the beginning because of the ``deterministic'' indices. Whenever the bad arm is observed, with high probability its UCB falls below the UCBs of the good arms, and hence, the bad arm is ``forgotten'' for some time, which leads to better initial performance.

\section{CONCLUSION}\label{sec:conc}

In this paper we analyzed the regret of CTS for CMAB with PTAs. We proved an order optimal regret bound when the expected reward function is Lipschitz continuous without assuming monotonicity of the expected reward function. Our bound includes the $1/p^*$ term that is unavoidable in general. Future work includes deriving regret bounds under more strict assumptions on the expected reward function such as the triggering probability modulated bounded smoothness condition given in \cite{wang2017improving} to get rid of the $1/p^*$ term. 
This is especially challenging due to the fact that the triggering probabilities depend on the expected arm outcomes, which 
makes relating the expected rewards under different realizations of $\bs{\theta}$ (such as given in the proof of Lemma  \ref{lemma:2}) difficult.

\subsection*{Acknowledgment}
This work was supported by the Scientific and Technological Research Council of Turkey (TUBITAK) under Grant 215E342.

\bibliographystyle{ieeetr}
\bibliography{OSA}

\onecolumn
\appendix

\section{SUPPLEMENTAL DOCUMENT}\label{sec:appendix}

\subsection{Additional Numerical Results}\label{sec:supp-numerical}
We provide additional results for the disjunctive version of the problem that is introduced in Section \ref{sec:numerical}. We consider the class of problems $B_{\mathrm{LB}}(R,K,p,\Delta)$ described in \cite{kveton_cascading}, where $L=1$ and the probability that user $1$ finds page $j$ attractive is given as
\begin{align*}
    p_{1,j}=\begin{dcases}
        p & \text{if } j\leq K \\
        p-\Delta & \text{otherwise.}
    \end{dcases} 
\end{align*}

Similar to \cite{kveton_cascading}, we set $p=0.2$ and vary other parameters, namely $R$, $K$, and $\Delta$. We run both CTS and CUCB for $100000$ rounds in all problem instances, and report their regrets averaged over $20$ runs in Table \ref{tbl:results}.

\begin{table}[!ht]
    \caption{Average regret and its standard deviation for CTS and CUCB for the class of problems $B_{\mathrm{LB}}(R,K,0.2,\Delta)$.}
    \label{tbl:results}
    \centering
    \begin{tabular}{rrr@{\hskip 0.25in}rrrr}
        \toprule
        & & & \multicolumn{2}{c}{\textbf{CTS}} & \multicolumn{2}{c}{\textbf{CUCB}} \\
        \cmidrule(lr){4-5} \cmidrule(lr){6-7}
        $R$ & $K$ & $\Delta$ & \textbf{Regret} & \textbf{Std. Dev.} & \textbf{Regret} & \textbf{Std. Dev.} \\
        \midrule
        16 & 2 & 0.15 & 155.4 & 14.1 & 1284.1 & 52.4 \\
        16 & 4 & 0.15 & 103.2 & 9.0 & 998.9 & 33.2 \\
        16 & 8 & 0.15 & 52.1 & 9.8 & 549.5 & 16.8 \\
        32 & 2 & 0.15 & 321.4 & 18.9 & 2718.8 & 61.2 \\
        32 & 4 & 0.15 & 252.2 & 17.0 & 2227.0 & 55.4 \\
        32 & 8 & 0.15 & 155.4 & 25.7 & 1531.0 & 21.9 \\
        16 & 2 & 0.075 & 276.9 & 50.7 & 2057.6 & 79.6 \\
        16 & 4 & 0.075 & 205.4 & 25.7 & 1496.5 & 65.2 \\
        16 & 8 & 0.075 & 113.1 & 40.4 & 719.4 & 53.7 \\
        \bottomrule
    \end{tabular}
\end{table}

We observe that CTS outperforms CUCB in all problem instances in terms of the average regret. Next, we compare the performance of CTS with CascadeKL-UCB proposed in \cite{kveton_cascading} using the results reported in Table 1 in \cite{kveton_cascading}. We observe that CTS outperforms CascadeKL-UCB in all problem instances as well. As a final remark, we see that for both CTS and CUCB, the regret increases as the number of pages ($R$) increases, decreases as the number of recommended items ($K$) increases, and increases as $\Delta$ decreases.

\subsection{Additional Facts}\label{sec:supp-facts}
We introduce Fact \ref{fact:1} in order to bound the expression in Lemma \ref{lemma:1} and Fact \ref{fact:3} in order to bound the expression in Lemma \ref{lemma:3}.

\begin{fact} \label{fact:1}
(Multiplicative Chernoff Bound (\cite{mitzenmacher2005probability} and \cite{chen}))	
Let $X_1,\ldots,X_n$ be Bernoulli random variables taking values in $\{0,1\}$ such that $\mathbb{E}[X_t|X_1,\ldots,X_{t-1}]\geq\mu$ for all $t \leq n$, and $Y = X_1+\ldots+X_n$. Then, for all $\delta \in (0,1)$,
\begin{align*}
\Pr[Y \leq (1-\delta)n\mu] \leq e^{-\frac{\delta^2n\mu}{2}} . 
\end{align*}
\end{fact}

\begin{fact} \label{fact:3}
(Results from Lemma 7 in \cite{wang18cts})
Given $Z \subseteq \tilde{S}^*$, let $\tau_j$ be the round at which $\mathcal{E}_{Z,1}(\pmb{\theta}(t)) \wedge \neg\mathcal{E}_{Z,2}(\pmb{\theta}(t))$ occurs for the $j$th time, and let $\tau_0=0$. If $\forall i\in Z, N_i(\tau_j+1)\geq q$ and $0 < \varepsilon\leq 1/\sqrt{e}$, then
\begin{align}
\mathbb{E}\left[ \sum_{t=\tau_j+1}^{\tau_{j+1}} \mathbb{I}\{\mathcal{E}_{Z,1}(\pmb{\theta}(t)),\mathcal{E}_{Z,2}(\pmb{\theta}(t))\} \right] \leq \prod_{i\in Z}B_q -1 \label{eqn:clmm7_1} 
\end{align}
where $B_q$ is given as
\rev{
\begin{align*}
B_q =
\begin{cases}
\min\left\{ \frac{4}{\varepsilon^2}, 1+6\alpha'_1\frac{1}{\varepsilon^2}e^{-\frac{\varepsilon^2}{2}q}+\frac{2}{e^{\frac{1}{8}\varepsilon^2q}-2} \right\} & \text{if } q>\frac{8}{\varepsilon^2} \\
\frac{4}{\varepsilon^2} & \text{otherwise}
\end{cases}
\end{align*}
}
and $\alpha'_1$ is a problem independent constant.

Moreover,
\begin{align}
\sum_{q=0}^{T} \left( \prod_{i\in Z} B_q-1 \right) \leq 13\alpha'_2 \left( \frac{2^{2|Z|+3}\log\frac{|Z|}{\varepsilon^2}}{\varepsilon^{2|Z|+2}} \right) \label{eqn:clmm7_2} 
\end{align}
where $\alpha'_2$ is a problem independent constant.
\end{fact}

\subsection{Proof of Lemma 1}\label{sec:lemma1}

The proof is similar to the proof of Lemma 3 in \cite{wang18cts}. However, additional steps are required to take probabilistic triggering into account. 
Consider a base arm $i \in [m]$. Let $\tau^i_w$ be the round for which base arm $i$ is in the triggering set of the selected super arm for the $w$th time. Hence, we have $i \in \tilde{S}(\tau^i_w)$ for all $w>0$. Also let $\tau^i_0=0$. Then, we have:
\begin{align}
    &\mathbb{E}[|\{ t:1\leq t\leq T, i\in\tilde{S}(t), |\hat{\mu}_i(t)-\mu_i|>\varepsilon \vee \mathcal{B}_{i,2}(t) \} |] \notag \\
    &\quad = \mathbb{E}\left[\sum_{t=1}^T \mathbb{I}\{i\in\tilde{S}(t), |\hat{\mu}_i(t)-\mu_i|>\varepsilon \vee \mathcal{B}_{i,2}(t)\} \right] \notag \\
    &\quad \leq \mathbb{E}\left[\sum_{w=0}^T \sum_{t=\tau^i_w+1}^{\tau^i_{w+1}} \mathbb{I}\{i\in\tilde{S}(t), |\hat{\mu}_i(t)-\mu_i|>\varepsilon \vee \mathcal{B}_{i,2}(t)\} \right] \notag \\
    &\quad = \sum_{w=0}^T \mathbb{E}[ \mathbb{I}\{ i\in\tilde{S}(\tau^i_{w+1}), |\hat{\mu}_i(\tau^i_{w+1})-\mu_i|>\varepsilon \vee \mathcal{B}_{i,2}(\tau^i_{w+1}) \}] \notag \\
    &\quad = \sum_{w=0}^T \mathbb{E}[ \mathbb{I}\{ |\hat{\mu}_i(\tau^i_{w+1})-\mu_i|>\varepsilon \vee \mathcal{B}_{i,2}(\tau^i_{w+1}) \}] \notag \\
    &\quad \leq 1 + \sum_{w=1}^T \Pr[ |\hat{\mu}_i(\tau^i_{w+1})-\mu_i|>\varepsilon \vee \mathcal{B}_{i,2}(\tau^i_{w+1}) ] \notag \\
    &\quad = 1 + \sum_{w=1}^T \Pr[|\hat{\mu}_i(\tau^i_{w+1})-\mu_i|>\varepsilon \wedge \neg\mathcal{B}_{i,2}(\tau^i_{w+1})] + \sum_{w=1}^T \Pr[\mathcal{B}_{i,2}(\tau^i_{w+1})] \notag \\
    &\quad = 1 + \sum_{w=1}^T \Pr[|\hat{\mu}_i(\tau^i_{w+1})-\mu_i|>\varepsilon, N_i(\tau^i_{w+1})>(1-\rho)wp_i] + \sum_{w=1}^T \Pr[N_i(\tau^i_{w+1})\leq (1-\rho)wp_i] \notag  \\
    &\quad \leq 1 + \sum_{w=1}^T \Pr[|\hat{\mu}_i(\tau^i_{w+1})-\mu_i|>\varepsilon, N_i(\tau^i_{w+1})>(1-\rho)wp^*] + \sum_{w=1}^T \Pr[N_i(\tau^i_{w+1})\leq (1-\rho)wp_i] 	\notag \\    
    &\quad \leq 1 + \sum_{w=1}^T 2 e^{-2 (1-\rho) w p^* \epsilon^2} + \mathbb{I}\{p^*<1\} \cdot \sum_{w=1}^T e^{-\frac{\rho^2wp^*}{2}} \label{eqn:hoeffding} \\
    &\quad \leq 1 + \frac{1}{(1-\rho)p^*\varepsilon^2} + \frac{2\mathbb{I}\{p^*<1\}}{\rho^2p^*} \notag
\end{align}
where the second term in \eqref{eqn:hoeffding} is obtained by observing that
\begin{align*}
&\Pr[|\hat{\mu}_i(\tau^i_{w+1})-\mu_i|>\varepsilon, N_i(\tau^i_{w+1})>(1-\rho)wp^*] \notag \\
&\leq \sum_{k = \lceil (1-\rho)wp^* \rceil}^{\infty} \Pr[|\hat{\mu}_i(\tau^i_{w+1})-\mu_i|>\varepsilon | N_i(\tau^i_{w+1}) = k] \Pr [N_i(\tau^i_{w+1}) = k]  
\end{align*}
and applying Hoeffding's inequality, and the third term in \eqref{eqn:hoeffding} is obtained by using Fact \ref{fact:1}.

\subsection{Proof of Lemma 2}\label{sec:lemma2}

The proof is similar to the proof of Lemma 1 in \cite{wang18cts}.
Let $\pmb{\theta'} := (\pmb{\theta'}_{\tilde{S}^*},\pmb{\theta}_{\tilde{S}^{*c}}(t))$ be such that 
\begin{align}
\|\pmb{\theta'}_{\tilde{S}^*}-\pmb{\mu}_{\tilde{S}^*}\|_{\infty}\leq\varepsilon ~. \label{eqn:closeness}
\end{align}

\textbf{Claim 1:} For all $S'$ such that $\tilde{S}'\cap\tilde{S}^*=\emptyset$, $S'\neq\text{Oracle}(\pmb{\theta'})$. 

Claim 1 holds since
\begin{align}
r(S',\pmb{\theta'}) &= r(S',\pmb{\theta}(t)) \label{eqn:L2_1} \\
&\leq r(S(t),\pmb{\theta}(t)) \label{eqn:L2_2} \\
&\leq r(S(t),\pmb{\mu}) + B \left( \frac{\Delta_{S(t)}}{B}-(\tilde{k}^{*2}+1)\varepsilon \right) \label{eqn:L2_3} \\
&= r(S(t),\pmb{\mu}) + \Delta_{S(t)} - B (\tilde{k}^{*2}+1)\varepsilon \nonumber \\
&= r(S^*,\pmb{\mu}) - B(\tilde{k}^{*2}+1)\varepsilon \label{eqn:L2_4} \\
&< r(S^*,\pmb{\mu}) - B\tilde{k}^*\varepsilon \nonumber \\
&\leq r(S^*,\pmb{\theta'}) \label{eqn:L2_5} 
\end{align}
where \eqref{eqn:L2_1} follows from Assumption \ref{a:rsmoothness} since $\pmb{\theta'}$ and $\pmb{\theta}(t)$ only differ on arms in $\tilde{S}^*$ and $\tilde{S}'\cap\tilde{S}^*=\emptyset$, \eqref{eqn:L2_2} holds since $S(t)\in\text{OPT}(\pmb{\theta}(t))$, \eqref{eqn:L2_3} is by $\neg\mathcal{D}(t)$ and Assumption \ref{a:rsmoothness}, \eqref{eqn:L2_4} is by the definition of $\Delta_{S(t)}$, and \eqref{eqn:L2_5} is again by Assumption \ref{a:rsmoothness}.

Next, we consider two cases:

\textbf{Case 1a:} $\tilde{S}^*\subseteq\tilde{\text{Oracle}}(\pmb{\theta'})$ for all $\pmb{\theta'} = (\pmb{\theta'}_{\tilde{S}^*},\pmb{\theta}_{\tilde{S}^{*c}}(t))$ that satisfies \eqref{eqn:closeness}. 

\textbf{Case 1b:} There exists $\pmb{\theta'} = (\pmb{\theta'}_{\tilde{S}^*},\pmb{\theta}_{\tilde{S}^{*c}}(t))$ that satisfies \eqref{eqn:closeness} for which $\tilde{S}^*\not\subseteq\tilde{\text{Oracle}}(\pmb{\theta'})$. For this $\pmb{\theta'}$, let $S_1 = \text{Oracle}(\pmb{\theta'})$ and $Z_1 = \tilde{S}_1\cap\tilde{S}^*$. Together with Claim 1, for this case, we have $Z_1\neq\tilde{S}^*$ and $Z_1\neq\emptyset$.

Note that Case 1a and Case 1b are complements of each other.

When Case 1a is true, for any given $\pmb{\theta'}$, with an abuse of notation, let $S_0 := \text{Oracle}(\pmb{\theta'})$. Then, we have $r(S_0,\pmb{\theta'}) \geq r(S^*,\pmb{\theta'}) \geq r(S^*,\pmb{\mu})-B\tilde{k}^*\varepsilon$. If $S_0\not\in\text{OPT}$, then we have $r(S^*,\pmb{\mu})=r(S_0,\pmb{\mu})+\Delta_{S_0}$. Combining the two results above, we obtain $r(S_0,\pmb{\theta'}) \geq r(S_0,\pmb{\mu}) + \Delta_{S_0} - B\tilde{k}^*\varepsilon$. By Assumption \ref{a:rsmoothness}, this implies that $\|\pmb{\theta'}_{\tilde{S}_0}-\pmb{\mu}_{\tilde{S}_0}\|_1 \geq \frac{\Delta_{S_0}}{B}-\tilde{k}^*\varepsilon > \frac{\Delta_{S_0}}{B}-(\tilde{k}^{*2}+1)\varepsilon$. Thus, from the discussion above, we conclude that either  $S_0\in\text{OPT}$ or $\|\pmb{\theta'}_{\tilde{S}_0}-\pmb{\mu}_{\tilde{S}_0}\|_1 > \frac{\Delta_{S_0}}{B}-(\tilde{k}^{*2}+1)\varepsilon$. This means $\mathcal{E}_{\tilde{S}^*,1}(\pmb{\theta'}) = \mathcal{E}_{\tilde{S}^*,1}(\pmb{\theta}(t))$ holds. Hence, if Case 1a is true, then Lemma \ref{lemma:2} holds for $Z = \tilde{S}^*$.

In Case 1b, we also have $r(S_1,\pmb{\theta'}) \geq r(S^*,\pmb{\theta'}) \geq r(S^*,\pmb{\mu})-B\tilde{k}^*\varepsilon$.
Consider any $\pmb{\theta''} = (\pmb{\theta''}_{Z_1},\pmb{\theta}_{Z_1^c}(t))$ such that 
\begin{align}
 \|\pmb{\theta''}_{Z_1}-\pmb{\mu}_{Z_1}\|_{\infty}\leq\varepsilon.  \label{eqn:closeness2}
 \end{align}

We see that 
\begin{align*}
\|\pmb{\theta''}_{\tilde{S}_1}-\pmb{\theta'}_{\tilde{S}_1}\|_1 &= \sum_{i\in\tilde{S_1}\cap\tilde{S}^*}|\theta''_i-\theta'_i| + \sum_{i\in\tilde{S_1}\cap\tilde{S}^{*c}}|\theta''_i-\theta'_i| \\
&\leq \sum_{i\in Z_1}(|\theta''_i-\mu_i|+|\mu_i-\theta'_i|) \\
&\leq 2(\tilde{k}^*-1)\varepsilon
\end{align*}
hence $r(S_1,\pmb{\theta''}) \geq r(S_1,\pmb{\theta'}) - 2B(\tilde{k}^*-1)\varepsilon \geq  r(S^*,\pmb{\mu})-B\tilde{k}^*\varepsilon-2B(\tilde{k}^*-1)\varepsilon = r(S^*,\pmb{\mu}) - B(3\tilde{k}^*-2)\varepsilon$.

\textbf{Claim 2:} For all $S'$ such that $\tilde{S}'\cap Z_1=\emptyset$, $S'\neq\text{Oracle}(\pmb{\theta''})$. 

Similar to Claim 1, Claim 2 holds since
\begin{align}
r(S',\pmb{\theta''}) &= r(S',\pmb{\theta}(t)) \nonumber \\
&\leq r(S(t),\pmb{\theta}(t)) \nonumber \\
&\leq r(S(t),\pmb{\mu}) + B \left(\frac{\Delta_{S(t)}}{B}-(\tilde{k}^{*2}+1)\varepsilon \right) \nonumber \\
&= r(S(t),\pmb{\mu}) + \Delta_{S(t)} - B(\tilde{k}^{*2}+1)\varepsilon \nonumber \\
&= r(S^*,\pmb{\mu}) - B(\tilde{k}^{*2}+1)\varepsilon \nonumber \\
&< r(S^*,\pmb{\mu}) - B(3\tilde{k}^*-2)\varepsilon \nonumber \\
&\leq r(S_1,\pmb{\theta''})  . \nonumber
\end{align}

Claim 2 implies that when Case 1b holds, we have $\tilde{\text{Oracle}}(\pmb{\theta''})\cap Z_1\neq\emptyset$. Hence, we consider two cases again for $\text{Oracle}(\pmb{\theta''})$:

\textbf{Case 2a:} $Z_1\subseteq\tilde{\text{Oracle}}(\pmb{\theta''})$ for all $\pmb{\theta''} = (\pmb{\theta''}_{Z_1},\pmb{\theta}_{Z_1^c}(t))$ that satisfies \eqref{eqn:closeness2}.

\textbf{Case 2b:} There exists $\pmb{\theta''} = (\pmb{\theta''}_{Z_1},\pmb{\theta}_{Z_1^c}(t))$ that satisfies \eqref{eqn:closeness2} for which $Z_1\not\subseteq\tilde{\text{Oracle}}(\pmb{\theta''})$. For this $\pmb{\theta''}$ let $S_2=\text{Oracle}(\pmb{\theta''})$ and $Z_2 = \tilde{S}_2 \cap Z_1$. Together with Claim 2, for this case, we have $Z_2\neq Z_1$ and $Z_2\neq\emptyset$.

Similar to Case 1a, when Case 2a is true, then Lemma 2 holds for $Z = Z_1$. Thus, we can keep repeating the same arguments iteratively, and the size of $Z_i$ will decrease by at least 1 at each iteration. After at most $\tilde{k}^*-1$ iterations, Case $(\cdot)$b will not be possible. In order to see this, suppose that we come to a point where $|Z_i|=1$. As in all iterations, either Case($i+1$)a or Case($i+1$)b must hold. However, when Case($i+1$)b holds, Claim $i+1$, which follows from Case($i$)b, implies that there exists a $Z_{i+1}\subseteq Z_i$ such that $Z_{i+1}\neq\emptyset$ and $Z_{i+1}\neq Z_i$, which is not possible when $|Z_i|=1$. Therefore, we conclude that some Case $(i+1)$a must hold, where $Z_i \subseteq\tilde{S}^*$, $Z_i \neq\emptyset$, and $\mathcal{E}_{Z_i,1}(\pmb{\theta}(t))$ occurs.

Finally, we need to show that Claim $i+1$ holds for all iterations. We focus on the claim
\rev{
\begin{align*}
r(S^*,\bs{\mu})-B(\tilde{k}^{*2}+1)\varepsilon &< r(S^*,\bs{\mu})-B(\tilde{k}^*+2\sum_{k=1}^{i}(\tilde{k}^*-k))\varepsilon
\end{align*}}
as repeating other arguments for all iterations is straightforward. The given inequality is true as \rev{$\tilde{k}^*+2\sum_{k=1}^{i}(\tilde{k}^*-k) \leq \tilde{k}^*+2\sum_{k=1}^{\tilde{k}^*-1}(\tilde{k}^*-k) = \tilde{k}^{*2} < \tilde{k}^{*2}+1$}. Note that, when checking Claim $i+1$, we know that $i$ previous iterations have passed, hence $\tilde{k}^*$ must be larger than $i+1$.

\subsection{Proof of Lemma 3}\label{sec:lemma3}
Given $Z$, we re-index the base arms in $Z$ such that $z_i$ represents $i$th base arm in $Z$. We also introduce a counter $c(t)$, and let $c(1)=1$. If at round $t$, $\mathcal{E}_{Z,1}(\pmb{\theta}(t)) \wedge \neg\mathcal{E}_{Z,2}(\pmb{\theta}(t))$ occurs and a feedback for $z_{c(t)}$ is observed, i.e., $z_{c(t)}\in S'(t)$, the counter is updated with probability $p^*/p_{z_{c(t)}}^{S(t)}$ in the following way:
\begin{align*}
c(t+1) =
\begin{cases}
c(t)+1 & \text{if } c(t)< |Z| \\
1 & \text{if } c(t)=|Z|
\end{cases}
\end{align*}
If the counter is not updated at round $t$, $c(t+1)=c(t)$. Note that when $\mathcal{E}_{Z,1}(\pmb{\theta}(t)) \wedge \neg\mathcal{E}_{Z,2}(\pmb{\theta}(t))$ occurs, $z_{c(t)}\in Z\subseteq \tilde{\text{Oracle}}(\pmb{\theta}(t))=\tilde{S}(t)$, hence we always have $0<p^*/p_{z_{c(t)}}^{S(t)}\leq 1$. Moreover, the probability that the counter is updated, i.e., $c(t+1)\neq c(t)$, given $\mathcal{E}_{Z,1}(\pmb{\theta}(t)) \wedge \neg\mathcal{E}_{Z,2}(\pmb{\theta}(t))$ occurs is constant and equal to $p^*$ for all rounds $t$ for which $\mathcal{E}_{Z,1}(\pmb{\theta}(t)) \wedge \neg\mathcal{E}_{Z,2}(\pmb{\theta}(t))$ occurs. To see this, consider a parameter vector $\pmb{\theta}$ such that $\mathcal{E}_{Z,1}(\pmb{\theta}) \wedge \neg\mathcal{E}_{Z,2}(\pmb{\theta})$ holds and let $S=\text{Oracle}(\pmb{\theta})$, then $\Pr[c(t+1)\neq c(t)|\pmb{\theta}(t)=\pmb{\theta}] = \Pr[z_{c(t)}\in S'(t)|S(t)=S]\cdot(p^*/p_{z_{c(t)}}^S) = p_{z_{c(t)}}^S\cdot(p^*/p_{z_{c(t)}}^S) = p^*$.

Let $\tau_j$ be the round at which $\mathcal{E}_{Z,1}(\pmb{\theta}(t)) \wedge \neg\mathcal{E}_{Z,2}(\pmb{\theta}(t))$ occurs for the $j$th time, and let $\tau_0 := 0$. Then, the counter is updated only at rounds $\tau_j$ with probability $p^*$. Let $\eta_{q,k}$ be the round $\tau_j$ such that $c(\tau_j+1)=k+1$ \revc{and $c(\tau_j)=k$} holds for the $(q+1)$th time. Let \revc{$\eta_{0,0} = 0$} and $\eta_{q,|Z|}=\eta_{q+1,0}$. We know that $0=\eta_{0,0}<\eta_{0,1}<\ldots<\eta_{0,|Z|}=\eta_{1,0}<\eta_{1,1}< \ldots$.

We use two important observations to continue with proof. Firstly, due to the way the counter is updated, for $t \geq \eta_{q,0}+1$ we have $N_i(t)\geq q$, $\forall i\in Z$.
Secondly, for non-negative integers $j_1$ and $j_2$, $\Pr[ \eta_{q,k+1} = \tau_{j_1+j_2+1} | \eta_{q,k} = \tau_{j_1} ]=p^*(1-p^*)^{j_2}$. This holds since for the given event to hold, the counter must not be updated at rounds $\tau_{j_1+1},\tau_{j_1+2},...,\tau_{j_1+j_2}$, each of which happens with probability $1-p^*$, and must be updated at round $\tau_{j_1+j_2+1}$ which happens with probability $p^*$.

Therefore, we have
\begin{align}
&\mathbb{E}\left[ \sum_{t=\eta_{q,k}+1}^{\eta_{q,k+1}} \mathbb{I}\{\mathcal{E}_{Z,1}(\pmb{\theta}(t)),\mathcal{E}_{Z,2}(\pmb{\theta}(t))\} \right] \notag \\
&\quad = \sum_{j_1=0}^\infty \Pr[ \eta_{q,k} = \tau_{j_1}] \sum_{j_2=0}^\infty \Pr[\eta_{q,k+1} = \tau_{j_1+j_2+1}| \eta_{q,k} = \tau_{j_1} ] \notag \\
&\quad\quad \times \sum_{j=j_1}^{j_1+j_2} \mathbb{E}\left[ \sum_{t=\tau_j+1}^{\tau_{j+1}} \mathbb{I}\{\mathcal{E}_{Z,1}(\pmb{\theta}(t)),\mathcal{E}_{Z,2}(\pmb{\theta}(t))\} \middle| \eta_{q,k}\leq\tau_j<\eta_{q+1,k} \right] \notag \\
&\quad \revc{\leq} \sum_{j_1=0}^\infty \Pr[\eta_{q,k} = \tau_{j_1}] \sum_{j_2=0}^\infty p^*(1-p^*)^{j_2} \sum_{j=j_1}^{j_1+j_2} \left(\prod_{i\in Z}B_q-1\right) \label{eqn:lmm3_1} \\
&\quad = \sum_{j_1=0}^\infty \Pr[ \eta_{q,k} = \tau_{j_1} ] \sum_{j_2=0}^\infty p^*(j_2+1)(1-p^*)^{j_2} \left(\prod_{i\in Z}B_q-1\right) \notag \\
&\quad = \sum_{j_1=0}^\infty \Pr[\eta_{q,k} = \tau_{j_1} ] \frac{1}{p^*}\left(\prod_{i\in Z}B_q-1\right) \notag \\
&\quad = \frac{1}{p^*}\left(\prod_{i\in Z}B_q-1\right) \notag
\end{align}
where \eqref{eqn:lmm3_1} holds due to our observations and \eqref{eqn:clmm7_1} in Fact \ref{fact:3}. 

Finally, we have
\begin{align}
\sum_{t=1}^T \mathbb{E}[\mathbb{I}\{\mathcal{E}_{Z,1}(\pmb{\theta}(t)),\mathcal{E}_{Z,2}(\pmb{\theta}(t))\}] &\leq \sum_{q=0}^{T} \sum_{k=0}^{|Z|-1} \mathbb{E}\left[ \sum_{t=\eta_{q,k}+1}^{\eta_{q,k+1}} \mathbb{I}\{\mathcal{E}_{Z,1}(\pmb{\theta}(t)),\mathcal{E}_{Z,2}(\pmb{\theta}(t))\} \right] \notag \\
&\leq \sum_{q=0}^{T} \sum_{k=0}^{|Z|-1} \frac{1}{p^*}\left(\prod_{i\in Z}B_q-1\right) \notag \\
&= \frac{|Z|}{p^*} \sum_{q=0}^{T} \left(\prod_{i\in Z}B_q-1\right) \notag \\
&\leq 13\alpha'_2 \cdot \frac{|Z|}{p^*} \cdot \left( \frac{2^{2|Z|+3}\log\frac{|Z|}{\varepsilon^2}}{\varepsilon^{2|Z|+2}} \right) \label{eqn:lmm3_2} 
\end{align}
where \eqref{eqn:lmm3_2} holds due to \eqref{eqn:clmm7_2} in Fact \ref{fact:3}.

\end{document}